\documentclass[sigconf]{acmart}

\usepackage{booktabs} 
\usepackage{graphics}
\usepackage{textcomp}
\AtBeginDocument{%
  \providecommand\BibTeX{{%
    \normalfont B\kern-0.5em{\scshape i\kern-0.25em b}\kern-0.8em\TeX}}}

\setcopyright{none}
\renewcommand\footnotetextcopyrightpermission[1]{}
\copyrightyear{2020} 
\acmYear{2020} 
\acmConference[Virtual, Augmented, and Mixed Reality for HRI]{HRI-20 Workshop on Virtual, Augmented, and Mixed Reality for Human-Robot Interaction}{March 23, 2020}{Cambridge, UK}

\begin{document}

\title{Implementing Virtual Reality for \\Teleoperation of a Humanoid Robot}

\author{Jordan Allspaw}
\affiliation{\institution{University of Massachusetts Lowell}}
\email{jallspaw@cs.uml.edu}

\author{Gregory LeMasurier}
\affiliation{\institution{University of Massachusetts Lowell}}
\email{gregory_lemasurier@student.uml.edu}

\author{Holly Yanco}
\affiliation{\institution{University of Massachusetts Lowell}}
\email{holly@cs.uml.edu}


\begin{abstract}
 Our research explores the potential of a humanoid robot for work in unpredictable environments, but controlling a humanoid robot remains a very difficult problem. 
 In our previous work, we designed a prototype virtual reality (VR) interface to allow an operator to command a humanoid robot. However, while usable, the initial interface was not sufficient for commanding the robot to perform the tasks; for example, in some cases, there was a lack of precision available for robot control. The interface was overly cumbersome in some areas as well. In this paper, we discuss numerous additions, inspired by traditional interfaces and virtual reality video games, to our prior implementation, providing additional ways to visualize and command a humanoid robot to perform difficult tasks within a virtual world.
\end{abstract}



\maketitle

\section{Introduction}
While significant research has been conducted with robots in domains such as telepresence, domestic assistance, and warehouse delivery, interaction methods for controlling humanoid robots are far less explored by comparison. The
largest exploration of the use of humanoid robots was conducted during the DARPA Robotics Challenge (DRC) where teams competed to perform tasks, such as opening a door, turning a valve, and walking up stairs \cite{darpaannounce}. An analysis of control methods and human-robot interaction (HRI) at the DRC found that teams used a variety of control methods including modifying individual joints, allowing for manual placement of footsteps, and setting waypoints to which the robot to plan and navigate \cite{DRC-Finals}.

One lesson learned by a DRC team was that full autonomy can be very time consuming to implement and adapt to new situations \cite{johnson2015team}. However, by using a shared control strategy, where some components are handled autonomously and some are handled by the human operator, the benefits of each can be maximized while reducing development time \cite{ferland2009egocentric}. 
Automated perception is an example of a task that is very difficult to work with in changing environments, yet tends to be trivial and quick for human operators with the right information. Even if the final goal is an autonomous solution, it can be desirable to start with a skilled operator first. With this in mind, we are first pursuing a shared control solution, where most of the decision making is performed by a skilled knowledgeable operator. The interface therefore needs to present the information and controls to allow the operator to perform their duties to a similar level to as if they were actually there.

We previously presented our initial implementations for a humanoid virtual reality interface \cite{allspaw2018teleoperating}. In that paper, we presented a limited set of controls and visualizations to facilitate teleoperation of a humanoid robot in virtual reality (VR). The primary focus for the interface was a humanoid bipedal robot; however, some parts of the system are applicable to a variety of other robot types. This initial interface allowed for an operator to command the robot using a set of high level commands. Through a limited set of interactions an operator could send goals to the robot and then approve or reject the plans. One of the main objectives was to create a fully complete VR interface. That is, we wanted to allow an operator to perform all of the tasks from within VR with at least similar ease and success as one could do with a traditional interface. With our initial proof of concept done, we decided to take another look at the previous interfaces developed for humanoid robots, as well as interfaces developed for video games from which we could draw inspiration.

Many VR video games have been released since our initial VR interface design. We have analyzed common control methods and interfaces found in these VR games for inspiration to upgrade our interface \cite{allspaw2019design}. We found that many of the new VR video games used similar controls as our prototype interface. Some of the similarities include using joysticks to control a character, interacting with objects using the controller's grip buttons, and giving the operator options to teleport around the VR world using a point and click method \cite{allspaw2019design}. These games also offered new design ideas which we have incorporated into our latest version of our interface. 

One new interface design includes the use of a "heads up display" (HUD) which would show information, such as health, above the operators wrist \cite{allspaw2019design}. This new type of interface has inspired our development of a wristwatch interface which we are using to display settings and other information about the robot's state.


In this paper we will discuss improvements to the visualization and control scheme for our new VR interfaces. The goal of our interface is to allow an operator to control a robot to perform dexterous tasks entirely from within VR, they should never need to remove the headset in order to use a command line or alternate interface. We also want to allow the operator to complete the tasks quickly and most importantly, accurately.

\section{Scenario}
Many related works have explored using VR in specific situations, such as haptic gloves for controlling a robot hand \cite{low2017hybrid} and the visualization of point clouds \cite{bruder2014poster}. VR was also used as a complementary interface during the DRC to observe task execution \cite{romay2017collaborative}. However, we are interested in a complete interface whereby an operator could control a humanoid bipedal robot entirely from within VR. Our VR interface is designed for controlling a semi-autonomous robot, capable of some autonomous tasks but not fully self-capable. This scenario is meant to encompass a semi-autonomous robot working in a very complex environment, which will therefore need supervision and oversight to carry out its tasks. 

For full VR control, the VR interface needs to include  means for controlling mobility, manipulation, and visualization of robot data. For mobility, the operator needs to be able to send high level navigation goals to the robot for its footstep planner. While the footstep planner is capable, there are many situations where the operator might be unsatisfied with the provided plan. In such a case, the interface needs to allow the operator to modify the plan by adjusting the individual footsteps or to cancel the plan. For manipulation, the operator needs to be able to send commands to the robot arm and end effectors. For visualization, the operator needs to be able to see and understand the sensor information which the robot can relay, in order to have an adequate situation and task awareness. 

We break down these capabilities into a series of tasks that the robot needs to be able to do: 
\begin{enumerate}
  \item Walk to a designated location, facing a specific direction;
  \item Avoid stepping on small objects while walking;
  \item Grab and pick up an object off a table;
  \item Manipulate an object such as turning a valve.
\end{enumerate}

Here we describe our designs for the interactions and visualizations that should allow the operator to control a robot in order to to complete these tasks accurately and in a timely manner. We are also interested in examining different ways of controlling the same action, to determine the best control method for each task.

\section{System Hardware Components}

For the front end we are using a HTC Vive \cite{niehorster2017accuracy} with the two included controllers. We also have an alternative setup with the same headset but substituting the controllers with the Manus VR Gloves \cite{manus-vr}. In both cases, there is position and orientation tracking of the operator's hands and head. The controllers provide the tracking natively while the Manus gloves are augmented with SteamVR Trackers that provide the feature. While the gloves do not have easy to use buttons provided on the controller, they add in accurate finger tracking and gesture control. In order to compensate for the loss of buttons the user can use a pinch to interact motion to select objects. We will also suggest several user interface (UI) elements to allow seamless control between the two modes.

The robot we are teleoperating is a Valkyrie R5 \cite{radford2015valkyrie} which is a bipedal humanoid robot created by NASA. It comes with a sophisticated balancing system that attempts to keep the robot standing upright while moving. The robot has two 7-dof arms, each with a 4 fingered hand, in addition to its 3-dof torso and 3-dof neck. It comes with a rgb-d sensor and lidar in the head, along with two rgb cameras in the torso, setup in a stereo configuration. In addition to the cameras and lidar it also has temperature sensors throughout it's joints, and has force sensors in the feet to detect irregular footsteps. All of this provides us with an abundant amount of information that needs to be carefully provided to the user to allow them to correctly assess the remote environment and robot state.

\section{User Interface (UI) Options}
One of the first improvements we determined to be necessary was the need for more UI elements, as well as a way to examine which elements should be used in which cases. In our analysis of comparable 2D interfaces, we identified several missing features in our system. 
While examining popular VR video games we found a large number of different visualization strategies and interaction methods, some of which would not carry over to robotics well, while ohters were promising candidates.  In order to organize our new design elements, we looked to Williams et al. \cite{UIMethods} who propose three principle categories for mixed-reality interaction design elements (MRIDEs), which are also applicable to elements in VR. Below we will discuss these principal categories as well as the UI elements we created that fall under each.

\subsection{Virtual Artifacts:}
Williams et al. define virtual artifacts as "3D objects that can be manipulated by either humans or robots (or which may move under their own ostensible volition), or which may impact the behaviors of robots" \cite{UIMethods}. We have designed two virtual artifacts for our VR system. 

The first virtual artifact in our VR system is a goal marker, commonly used in standard 2D interfaces. When the operator sets a goal to which the robot should navigate, an object is spawned at the target location in the virtual world. The operator can then see the virtual marker in relation to other VR elements while moving around their virtual avatar. The goal marker could improve the operator's situational awareness because they are able to look out into the virtual world and see to where the robot is planning to navigate. 

The next virtual artifacts that we use are footstep markers, as seen in \ref{fig:footstep}. These footstep markers are created by a footstep planner when the operator sends a goal marker, and are rendered into the VR world. By viewing the footsteps in the VR world relative to other elements, the operator can evaluate whether the planned footsteps are satisfactory. This method can potentially improve the operator's situational awareness as they are able to view the exact path, including every single footstep the robot plans take, to navigate to a goal. The operator is able to interact with and move each footstep marker which will then update the corresponding footstep in the footstep plan. An operator may want to adjust the robot's planned footsteps to make them wider apart, to compensate for sensor inaccuracies, or to avoid obstacles that were not avoided by the planner.

\subsection{User-Anchored Interface Elements:}
User-anchored interface elements are "anchored to points in the user’s camera’s coordinate system" \cite{UIMethods}. We have implemented two user-anchored interface elements into our system design. 

\begin{figure}
\includegraphics[width=\columnwidth]{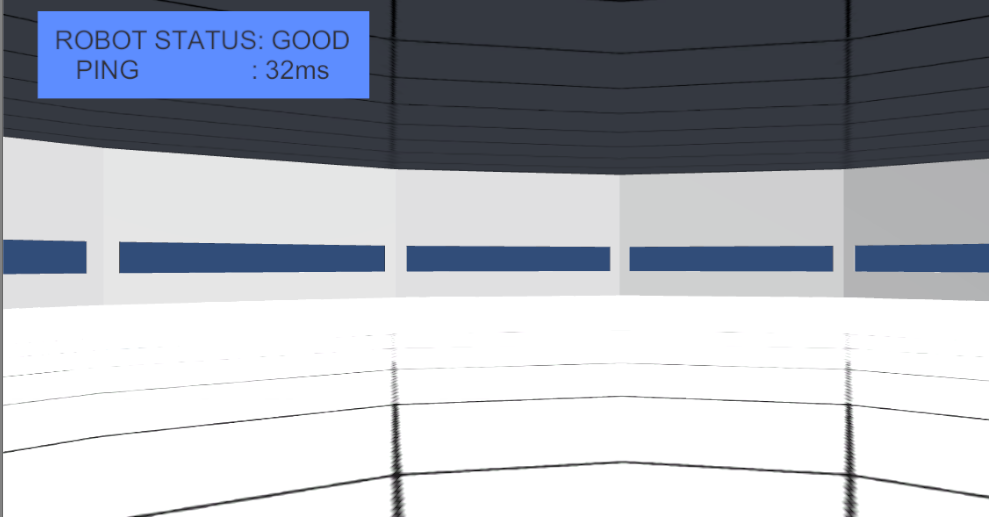}
  \caption{HUD attached to the users camera view, on the top left corner of their vision}
\label{fig:statushud}
\end{figure}

The first user-anchored interface element we added was a heads-up display (HUD), something commonly seen in traditional interfaces. This HUD is attached to the user camera view at the top and corners of the user's vision. An observation we had was that putting too much information permanently attached to the user's view can be distracting, so this should be used sparingly. UI elements can be hidden or faded on the HUD, then brought into view when important, such as low network status or a low battery. This HUD is anchored at the top left of the operator's view, as seen in \ref{fig:statushud}. 

\begin{figure}
\includegraphics[width=\columnwidth]{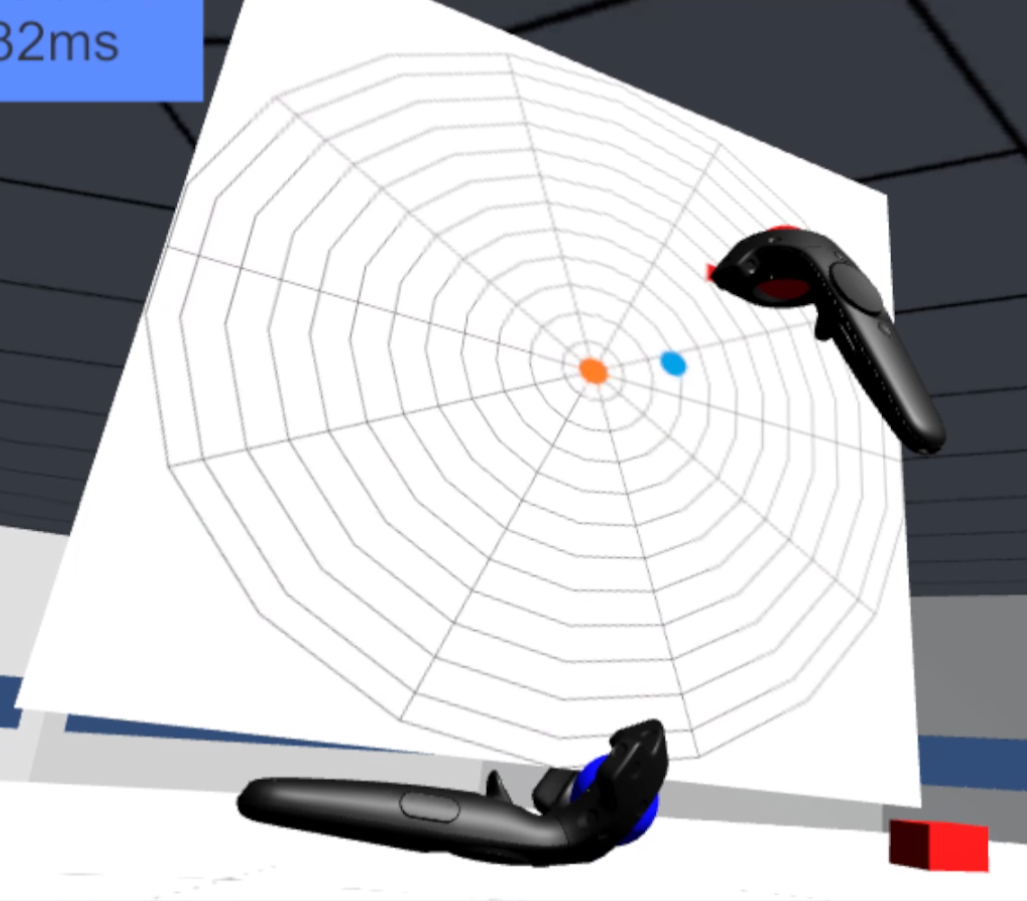}
  \caption{Minimap brought up on the wristwatch element. The orange circle in the center is the robot, the blue circle is the operator's avatar, and the red box is the goal location to which the robot was told to navigate. In this case, the operator tapped the minimap to indicate to where the robot should walk.}
\label{fig:minimap}
\end{figure}

Another addition primarily inspired from video games is the idea of a virtual wristwatch. As the operator holds their wrist up, as if examining a watch on their wrist, a UI element will appear. This window moves with the operator's hand and will automatically hide if the operator moves their hand away. The virtual wristwatch is great for displaying information that an operator may want to quickly access but is not necessary to be viewed at all times. Our virtual wristwatch can be used to display information including settings, robot status, battery levels, joint control \ref{fig:wristui}, camera streams, and an overhead view of the virtual world \ref{fig:minimap}.

\subsection{Environment-Anchored Interface Elements:}
Environment-anchored interface elements are "anchored to points in the coordinate system of a robot or some other element of the environment" \cite{UIMethods}. 

The first example of an environment anchored interface element we created was rendering the point cloud. Since the robot is equipped with a RGB-D sensor, we are able to scan the environment and add that point cloud, anchored to the tf tree of the robot. In this way, we are able to visualize the point cloud in the VR world. 

\begin{figure}
\includegraphics[width=\columnwidth]{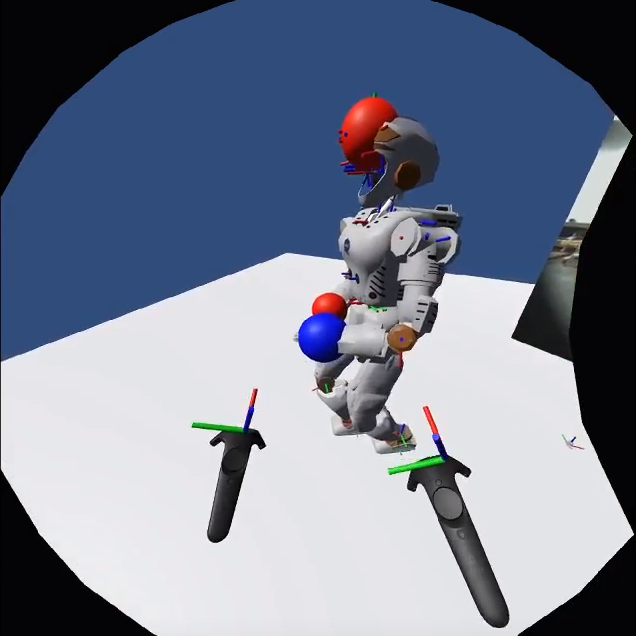}
  \caption{Robot Model being displayed in VR, with the positions and joint states updated in real time. The balls located on the hand and end are interaction markers; the operator can grab them and drag them to move their respective robot component.}
\label{fig:valvr}
\end{figure}

Another environment-anchored element was is the robot model rendered in the VR world which is updated in real time as seen in figure \ref{fig:valvr}.

Our VR system also makes use of a virtual tablet as an environment-anchored interface element. We have expanded the functionality of the virtual tablets which were part of our initial system design \cite{allspaw2018teleoperating}. Now, operators can either hold or place the virtual tablets in the virtual world. Like the wristwatch, these tablets can contain information, such as a camera stream. The primary advantage of these tablets is that an operator can have several of them at once, and they can be placed in an orientation that the operator finds useful. The downside is that unless the operator directly moves them, they remain fixed in the virtual world. If the robot travels to a different area in the world, the operator would need to move these elements as well. Given this limitation, these tablets are primarily useful if the robot is working in a fixed workspace for a period of time.

\begin{figure}
\includegraphics[width=\columnwidth]{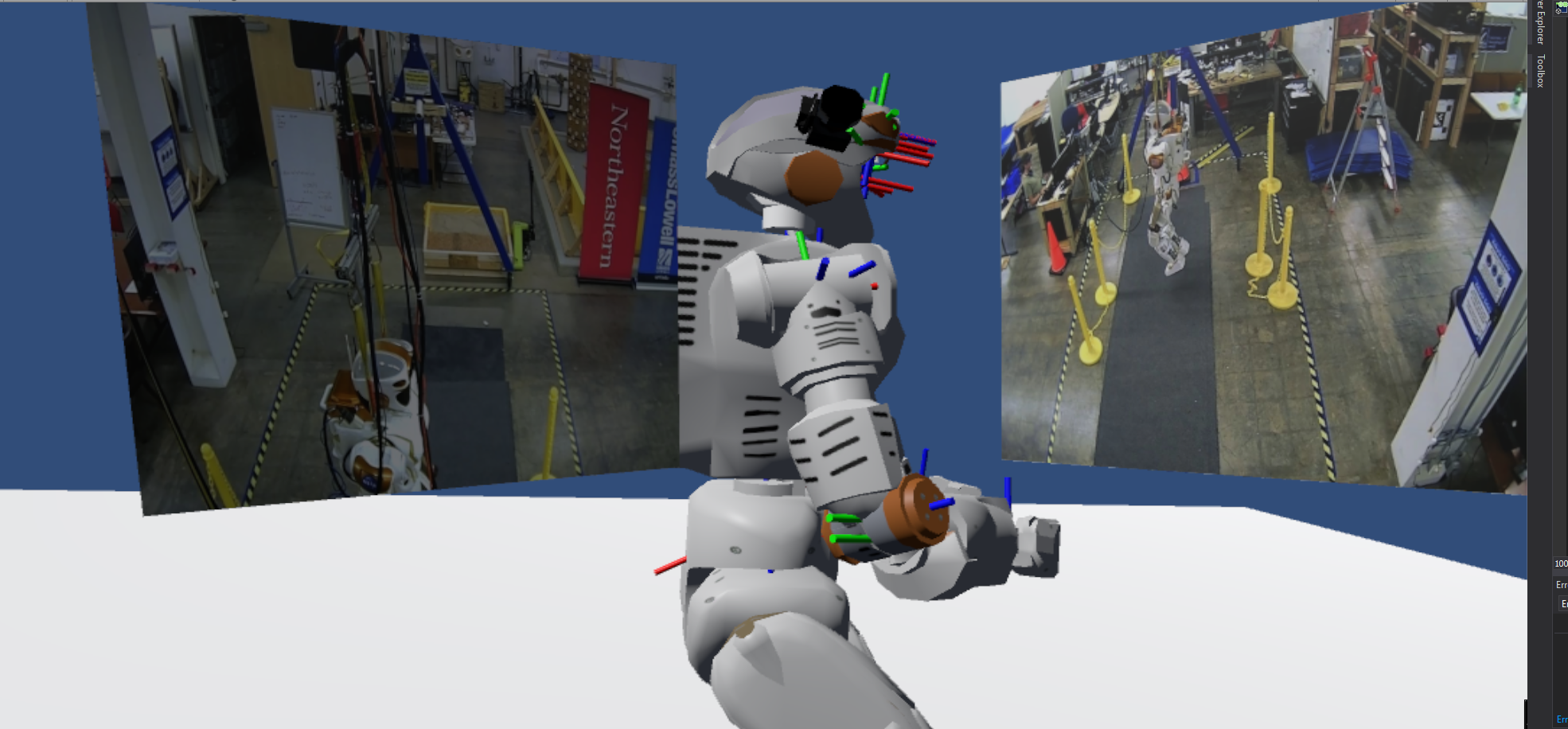}
  \caption{Camera views streamed to a virtual tablet. The tablet views can be placed and moved in the environment in locations that the operator desires.}
\label{fig:cameraviews}
\end{figure}


\section{Mobility}
 
 
 When a robot generates a footstep plan, it is not guaranteed that every footstep is correct, due to errors such as inaccurate sensors which can result in incorrect height maps \cite{IHMC-DRC}. To overcome this issue several teams in the DRC had the ability to manually modify and add footsteps to their plan \cite{DRC-Finals}. One team's prior approach to this type of interface was to allow a human operator to confirm and modify the position of regions, which were generated by their footstep planning algorithm, where footsteps are possible \cite{MIT-footstep}. This team only checked to confirm that the robot ended its step in a collision free state; however, they did not check for collisions while the robot was transitioning to its next step \cite{MIT-footstep}. 
 Here, VR would really be beneficial to the operator. An operator would be able to monitor the robot in real time, with an increased situation awareness, allowing operators to prevent some of these collisions. Another team developed a footstep planning algorithm that increased their success rate by using partial footholds \cite{IHMC-Algorithm}. They also created an interface which would notify the operator if a footstep was out of reach \cite{IHMC-DRC}. To compensate for footstep errors they designed an interface where their operators could adjust each footstep \cite{IHMC-DRC}. After the operator modified a footstep, an algorithm would snap the footstep onto the terrain \cite{IHMC-DRC}. This is one capability our prior interface lacked; the operator could approve and decline a footstep plan, but short of just re-planning over and over the operator could not change the plan directly.

\begin{figure}
\includegraphics[width=\columnwidth]{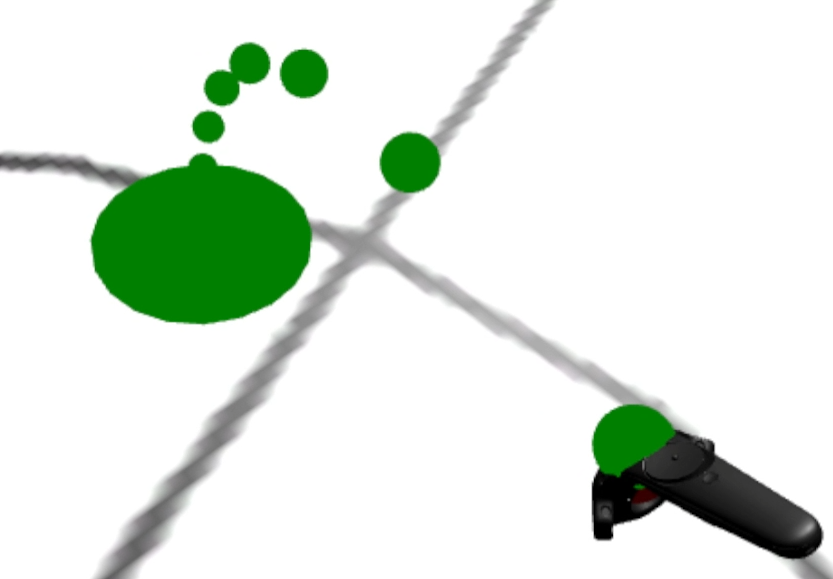}
  \caption{Pointer from the VR controller. The small dots show the trajectory for depth perception, while the large circle aligns with the floor to where the operator is pointing. The trail is arced for easier fine control rather than a direct laser pointer style.}
\label{fig:pointer}
\end{figure}
 
 The first mobility method is using a joystick on the handheld controller, similar to how one would use a joystick on any handheld controller. Such a device has commonly been used to control mobile devices ranging from generic radio-controlled (RC) toy cars to emergency response robots such as the Packbot \cite{yamauchi2004packbot}. Pressing forward on the joystick will instruct the robot to walk forward until the joystick is released. This control scheme is easily utilized on the controller; however, this interface is not as intuitiv on gloves becausee they do not have built in joysticks. This scheme is intended to be real time and less precise.

\begin{figure}[t]
 \centering
\includegraphics[width=\columnwidth]{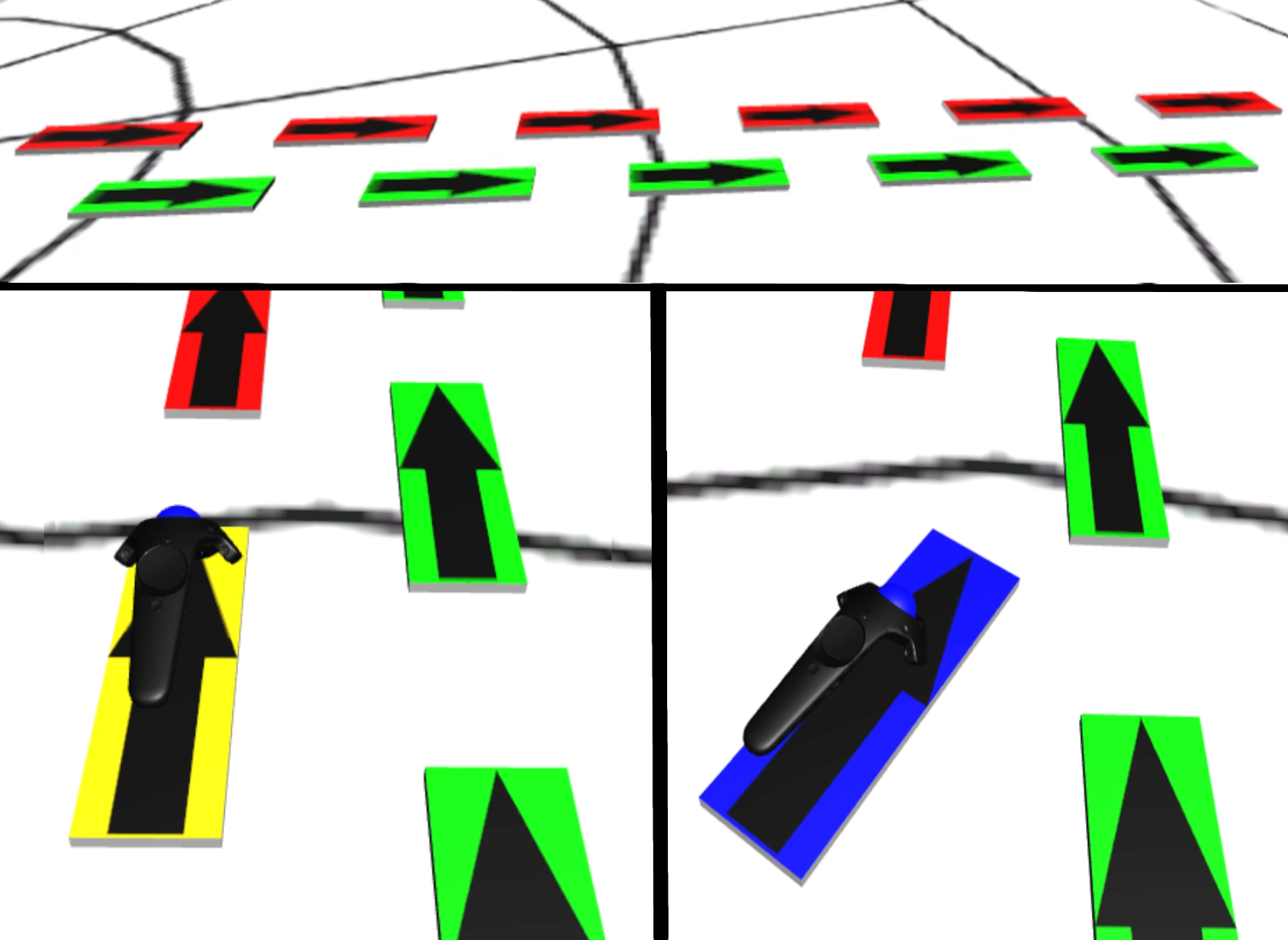}
  \caption{Examples of footsteps created by the footstep planner. The top image shows the footsteps from a side view, with red markers for the left foot and green markers for the right foot. The bottom left and right displays show the operator hovering over a footstep highlighting it yellow, then pressing a button turning the marker blue indicating the operator can move the controller to modify the footstep. The operator can rotate and move the footsteps along the floor until they are happy with the new plan and confirm the changes. }
\label{fig:footstep}
\end{figure}

 The second method of navigation is by pointing to a location in the VR world. Pressing down a button on the joystick brings up an arc pointer towards the location. Releasing the button will send that destination as a goal to the footstep planner, as seen in ref \ref{fig:pointer}. The footstep planner then publishes a list of the footsteps that will then be visualized in the VR world. The operator can view these footsteps, and then modify them if desired as seen in \ref{fig:footstep}. For example, the operator can make the feet slightly wider apart or sidestep an obstacle the planner failed to avoid. When the operator is satisfied with the footsteps, they confirm the plan and send it to the robot. This method is not necessarily real time and requires operator confirmation. 
 
 Navigating around the virtual world with a 1-to-1 scale avatar can be very useful for keeping the operator's obstacle awareness high; however it can be cumbersome to move around constantly. The third control scheme proposed allows the operator to bring up a virtual intractable minimap inspired interface. The operator is able to point to a location on the virtual tablet, and that goal will appear in the virtual world as a destination goal. The footstep planner would then plan a path to the goal and the operator could approve or decline the steps as before. An example is shown in \ref{fig:footstep}.
 
\section{Joint Control}
Much like footsteps, joint positions can also be inaccurate due to inaccurate sensors. One way to compensate for joint control inaccuracies is to allow the operator to control each joint. At the DRC trials, every team had an interface to control each joint individually \cite{DRC-Trials}. Teams also used this type of control method during the DRC finals \cite{DRC-Finals}. Our prior interface was missing a similar, more direct, joint control method. Another control method that teams used to compensate for inaccuracies was to move the joints on a 3D model using a Cartesian transform tool \cite{DRC-Trials} \cite{DRC-Finals}. Our prior interface allowed operators to modify a robot's joints in a similar manner, by grabbing and dragging the joints on a 3D model of the robot. 

While walking can remain at a high level where the robot can reliably handle footsteps, there is a need to provide finer control methods for the robot's head, torso, two 7-dof hands, and the fingers attached to each hand. Depending on the task, different control methods may be desirable so we have prepared several. 

The first is by enabling a direct control mode, whereby the robot would mimic the operator's actions in the VR world in a 1-to-1 ratio. Both the gloves and controllers allow the operator to move their own hand around, which will then send their position and use inverse kinematics to move the robot arm. This method is a very quick and easy way to get the robots hand to move to a desired 3D location; however, it does not allow for much customization. For example, the operator's elbow position and orientation is not tracked, so the elbow joints of the robot will be controlled by the inverse kinematics (IK) solver, not by the operator. 

\begin{figure}
\includegraphics[width=\columnwidth]{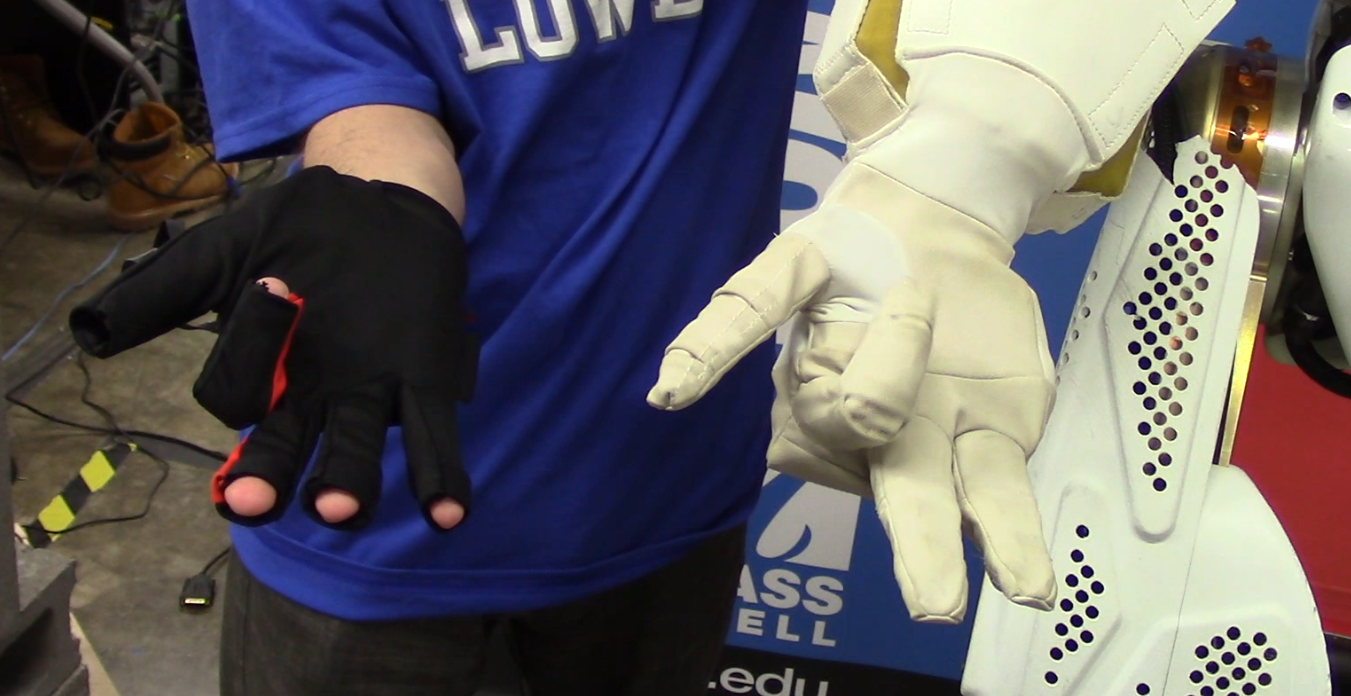}
  \caption{Manus VR Glove being used to control the robot's fingers in a direct 1-to-1 mode.}
\label{fig:manus1to1}
\end{figure}

The glove has an additional level of control because in addition to the hand position and orientation being tracked; it can also track the fingers, something the controller cannot do. An example can be seen in \ref{fig:manus1to1}. An alternate form of this control is rather than mapping motion 1-to-1, allowing the operator to simply grab the virtual robot hand and move it where they want the robot to move, an example of which can be seen in \ref{fig:valvr} This alternate style provides no new functionality, but may prove more intuitive than toggling between modes. It also allows the operator to view the plan and approve it before the plan is executed. Both of these styles involve the use of inverse kinematics, which while convenient can occasionally lead to irregular arm movements. As such we will also discuss more direct methods of control.

 \begin{figure}
\includegraphics[width=\columnwidth]{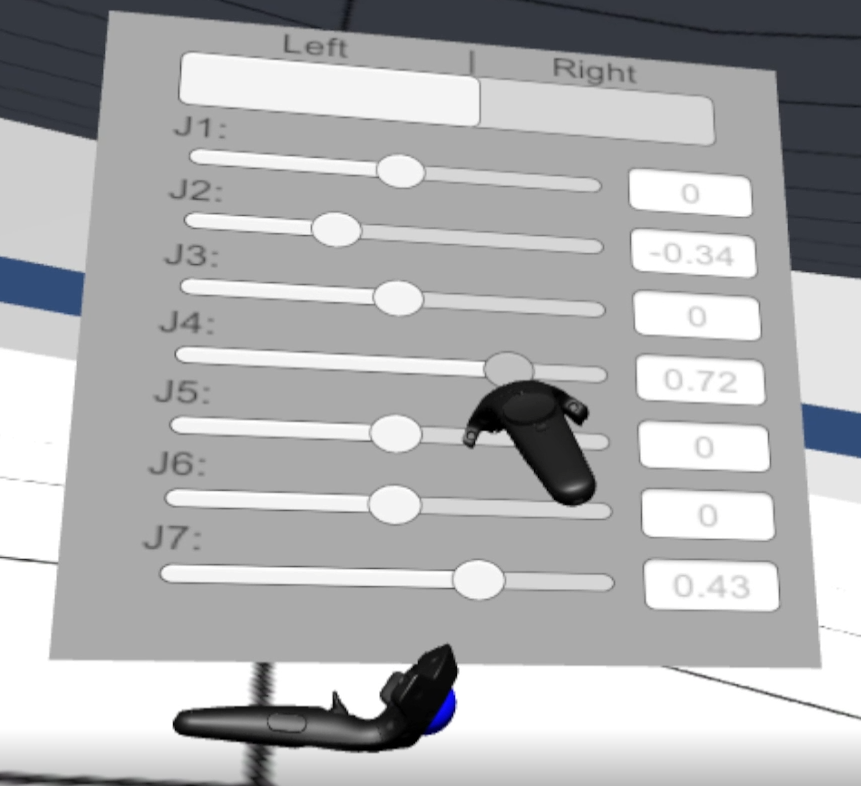}
  \caption{Example of the direct joint control displayed to the operator's wristwatch. As a fallback, the operator can grab and drag the sliders for the joints, with J1 corresponding to the shoulder pitch joint etc.}
\label{fig:wristui}
\end{figure}

The next level of control is the ability to grab specific joints and move them. This is a type of control commonly found in frameworks such as Moveit \cite{chitta2012moveit}. The advantage in VR is that an operator can easily see the robot's 3D model, as well as the nearby environment. This method is viable for both the glove and the controller. Here the user can simply grab both sides of a joint in the robot arm and turn it, causing the joint to change. This allows the operator to change one joint at a time in a very direct manner. 

The final level of control is a fallback which provides sliders inside the virtual world. The operator can use either the wristwatch UI or a virtual tablet UI, in order to manipulate the sliders which directly control joints on the robot, an example of which can be seen in figure \ref{fig:wristui}. To keep the UI small, the operator can choose which joint to control (i.e., left arm, right arm etc.). Then they can change the sliders that correspond to the robot's motion. This fallback is for scenarios where direct control is necessary and all other forms of control are inadequate, such as correcting an inaccurate IK goal position. 

\section{Discussion}
Our primary focus for this VR interface was a humanoid bipedal robot; however, many of the control methods we have designed can be generalized to other types of robots, such as industrial or mobile robots.

Our joystick control method can be generalized to both mobile robots and industrial robots. A joystick interface is commonly used to teleoperate RC cars and mobile robots, and robots such as Fetch Robotics' Fetch \cite{fetch} and iRobot's Packbot \cite{yamauchi2004packbot} can be teleoperated using an included controller. Other work has investigated using a joystick interface for mobile robot mobility \cite{vrMobileRobot}. This control method would not be the best option for controlling an industrial robot because industrial robots have many joints and they are often moving around in 3D space. The joystick could be used to control the position of an end effector, for example, but they will not be able to control the position and orientation of each individual joint. Other work has investigated using a combination of two joysticks to control an industrial robot arm \cite{armJoystick}. Some mobile robots, such as drones, also have to navigate in 3D space and they are often operated with similar interfaces which use a joystick interface in combination with other joysticks or buttons.

A point and click navigation method can be applied to both mobile robots and industrial robots. An operator can control a mobile robot, much like a bipedal humanoid robot, where they point to a location to which they want the robot to navigate, then the robot plans to navigate to the specified goal. A mobile robot's navigation plan may require adjustment due to inaccurate sensors, so an interface similar to our footstep markers could be used to modify the robot's planned path. Industrial robots can also be controlled using this point and click method. An operator could point at a location which the robot could then pass into its IK solver as the goal position. A similar interface has been used in the real world where an operator uses laser pointer to command a robot arm to grab a particular object in a scene \cite{laser-pointer}.

Our minimap interface can also be used to control both mobile robots and industrial robots. Mobile robots can be controlled using a minimap in the same way that we use to control a bipedal humanoid robot. The operator can tap on the minimap which will generate a goal for the robot to navigate to. This can be generalized to industrial robots by using an map of the overhead view of the workspace. An operator could then tap to a location on the workspace which can be used as a goal for the robot's IK solver. If an operator were to only use this method, they would not be able to control each individual joint and they will also not have any control over the height of the goal as the map would only give 2D position.

The 1-to-1 joint control method can be generalized to industrial robots, but would not be particularly useful for mobile robots. This control method is best when the robot has the same number of joints as a human. An operator would not be able to intuitively control a robot with more joints because the operator can not physically bend their arms in the ways that many industrial robots can. Other work has investigated using VR to directly control industrial robots \cite{vrBaxter}. This method can be used to give the robot a target goal for its IK solver, but the operator will not be able to directly control every joint using this method.

Our grab and drag control method is applicable to industrial robots as well. An operator will be able to grab and drag the joints of an industrial robot, in the same way that they control the joints on a bipedal humanoid robot. They can grab each joint and adjust its position and orientation.

The final control method we discussed, using sliders for joint control, is also applicable to industrial robots. An operator can control the orientation and position of each joint by moving the slider interface element in the same way that was used to control the joints on a bipedal humanoid robot.
    
\section{Future Work}
So far we have discussed many different options for HRI in VR. We focused on methods to provide a large number of capabilities to an operator in virtual reality without overloading them with too many buttons or modes. One way to add more interaction methods without adding more visual elements is to add some amount of voice control to the system. While there has been a significant amount of research in natural language processing for human-robot instruction, it would be interesting to combine it with virtual reality to see if the two domains could augment each other. 

Another component we are interested in expanding is the ability to view planned commands being carried out before they occur. For example, viewing a simulated robot perform the commands before approving for the final plan. This could allow the operator to more accurately curate commands sent.

We plan to conduct an evaluation of our interfaces. We will conduct a user study to examine these two VR interfaces, compared to a standard 2D interface, which was modeled heavily after interfaces commonly used during the DRC. Our goal is to verify that the provided types of controls and visualizations are sufficient to control a humanoid robot to perform complex tasks, such as those seen in the DRC, at a comparable level to other interfaces used. We are also interested in investigating weaknesses of our interfaces, whether it be in unnecessary or redundant features, or in lacking capability.

Another avenue to explore is what changes would exist when using this interface for different types of robots, for example, a more common robot with a single end effector. While many of the design decisions were made specifically for a humanoid robot, much of the interface would be highly adaptable to more typical robot styles. 

\section{Conclusion}
In this paper, we discussed improvements we have made to our initial VR interface. After examining other interfaces as well as similar VR products in gaming, we came up with a number of additions and improvements to our interface. 

In order to provide more control in situations where the autonomous footstep planner fails to provide safe footsteps, we created a goal marker and movable footstep markers. The footstep markers allow the operator to adjust and modify the footstep path to compensate for sensor and planner inaccuracies. This brings the footstep planning and interaction in line with comparable 2D interfaces. 

In order to address the limited number of visualizations and interactions available we have also added a HUD, wristwatch interface, and virtual tablets to display information including settings, robot state information, camera streams, and joint control. The virtual tablets are a reclassification of something we were already using, but expanded for additional options. The HUD was inspired primarily from video games as a way of putting important information within the operator's view at all times, a primary example being network status to the robot. The wristwatch was a new idea also inspired from gaming, which served multiple functions including allowing us to offload functionality from specific buttons on the controller and providing another option for visualization. 

We also discussed our methods for mobility including using a joystick on the controllers, a point and click method, and a minimap interface. Finally for controlling the robots arms, neck and torso we proposed control methods including a 1-to-1 mimicking method, a method where the operator can grab and move the robot's virtual model. 

Our goal is to allow the operator to teleoperate a robot to perform complex tasks. While many of the interaction methods revolve around various semi-autonomous controls, sometimes the robot will just not be able to plan for a situation and a fallback is required. To address this we discussed a slider control method to provide direct joint control, for when other solutions involving inverse kinematics are undesirable. This still allows the operator to retain the situation and task awareness provided by VR, but allows the operator to fine tune motions in cases where careful precision is required. 

All of these visualizations and controls should allow an operator with a small amount of training to complete tasks in VR to a comparable or superior level to those by other traditional robot interfaces.

\begin{acks}
This work has been supported in part by the National Science Foundation (IIS-1944584 and IIS-1451427), the Department of Energy (DE-EM0004482), and NASA (NNX16AC48A).
\end{acks}

\bibliographystyle{ACM-Reference-Format}
\bibliography{vr-ref}

\end{document}